%% file: egpaper_final.tex
\documentclass[10pt,twocolumn,letterpaper]{article}
\usepackage{float}
\usepackage{comment}
\usepackage{caption}
\usepackage{stfloats}
\usepackage{amsfonts}
\usepackage[toc,page]{appendix}

\usepackage{cvpr}
\usepackage{times}
\usepackage{epsfig}
\usepackage{graphicx}
\usepackage{amsmath}
\usepackage{amssymb}
\usepackage{mathptmx}
\usepackage{mathrsfs}
\usepackage{comment}
\usepackage{authblk}
\usepackage{wrapfig}
\usepackage[pagebackref=true,breaklinks=true,letterpaper=true,colorlinks,bookmarks=false]{hyperref}
\def\cvprPaperID{3}

\cvprfinalcopy
\ifcvprfinal\fi
\begin{document}

\title{Adversarial Texture Optimization from RGB-D Scans}

\makeatletter
\renewcommand\Authfont{\fontsize{11.5}{14.4}\selectfont}
\renewcommand\AB@affilsepx{\qquad \protect\Affilfont}
\makeatother
\author[1,3]{Jingwei Huang}
\author[2]{Justus Thies}
\author[2]{Angela Dai}
\author[3]{Abhijit Kundu}
\author[3,4]{Chiyu ``Max'' Jiang}
\author[1]{Leonidas Guibas}
\author[2]{\\ Matthias Nie{\ss}ner}
\author[3]{Thomas Funkhouser}
\affil[1]{Stanford University}
\affil[2]{Technical University of Munich}
\affil[3]{Google Research}
\affil[4]{UC Berkeley}

\twocolumn[{%
\renewcommand\twocolumn[1][]{#1}%
\maketitle
\vspace{-0.3in}
\begin{center}
    \centering
    \includegraphics[width=\textwidth]{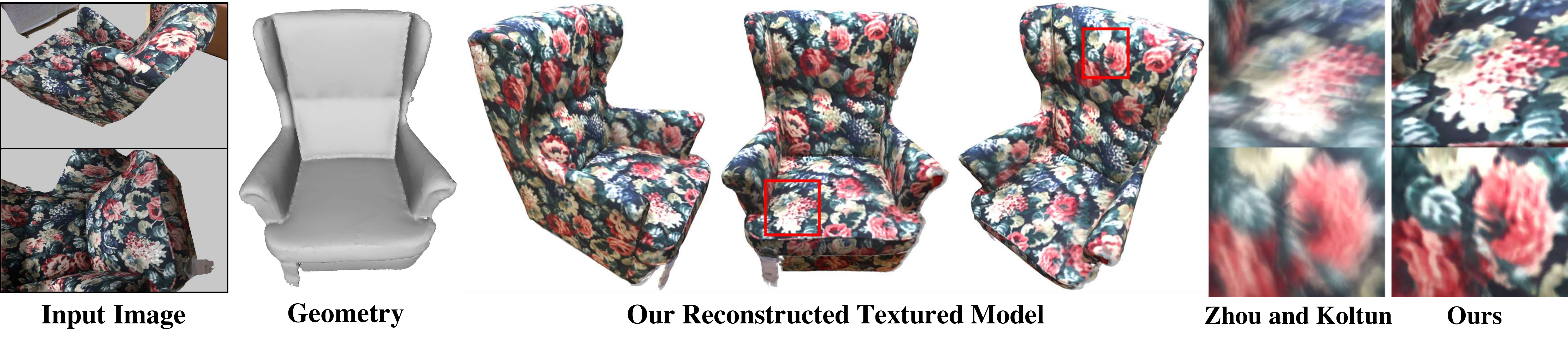}
    \captionof{figure}{Our goal is to reconstruct high-quality textures from an RGB-D scan. Unlike traditional methods which optimize for a parametric color map to reduce misalignment error (Zhou and Koltun~\cite{zhou2014color}), we learn a misalignment-tolerant discriminator, producing sharper textures.}
    \label{fig:teaser}
\end{center}%
}]

\maketitle

\begin{abstract}
   Realistic color texture generation is an important step in RGB-D surface reconstruction, but remains challenging in practice due to inaccuracies in reconstructed geometry, misaligned camera poses, and view-dependent imaging artifacts.
   In this work, we present a novel approach for color texture generation using a conditional adversarial loss obtained from weakly-supervised views.
   Specifically, we propose an approach to produce photorealistic textures for approximate surfaces, even from misaligned images, by learning an objective function that is robust to these errors.
   The key idea of our approach is to learn a patch-based conditional discriminator which guides the texture optimization to be tolerant to misalignments.
   Our discriminator takes a synthesized view and a real image, and evaluates whether the synthesized one is realistic, under a broadened definition of realism.
   We train the discriminator by providing as `real' examples pairs of input views and their misaligned versions -- so that the learned adversarial loss will tolerate errors from the scans.
   Experiments on synthetic and real data under quantitative or qualitative evaluation demonstrate the advantage of our approach in comparison to  state of the art (see Figure~\ref{fig:teaser}, right). Our code is publicly available\footnote{\url{https://github.com/hjwdzh/AdversarialTexture}} with video demonstration\footnote{\url{https://youtu.be/52xlRn0ESek}}.\vfill
\end{abstract}

\input{0Introduction}
\input{1RelatedWorks}
\input{2Approach}

\input{3Experiments}

\input{4Result}

\input{5Conclusion}

\section*{Acknowledgements}
This work was supported by the ZD.B and ERC Starting Grant \textit{Scan2CAD} (804724), NSF grants  CHS-1528025 and IIS-1763268, a Vannevar Bush Faculty Fellowship, and grants from the Samsung GRO program and the Stanford SAIL Toyota Research Center.

{\small
\bibliographystyle{ieee_fullname}
\bibliography{egbib}
}

\input{6supplemental}
\end{document}

%% file: 0Introduction.tex
\vspace{-1.3cm}
\section{Introduction}

The wide availability of consumer range cameras has spurred extensive research in geometric reconstruction of real-world objects and scenes, with state-of-the-art 3D reconstruction approaches now providing robust camera tracking and 3D surface reconstruction~\cite{newcombe2011kinectfusion,izadi2011kinectfusion,whelan2015elasticfusion,dai2017bundlefusion}.
However, producing photorealistic models of real-world environments requires not only geometric reconstruction but also high-quality color texturing.
Unfortunately, due to noisy input data, poorly estimated surface geometry, misaligned camera poses, unmodeled optical distortions, and view-dependent lighting effects,  aggregating multiple real-world images into high-quality, realistic surface textures is still a challenging problem.
In order to overcome these problems, various approaches have been developed to optimize color textures using models to adjust camera poses~\cite{zhou2014color,huang20173dlite}, distort images~\cite{bi2017patch,huang20173dlite,zhou2014color}, and balance colors \cite{huang20173dlite,zhou2014color}.  However, these prior approaches are not expressive enough and/or their optimization algorithms are not robust enough to handle the complex distortions and misalignments commonly found in scans with commodity cameras -- and therefore they fail to produce high-quality results for typical scans, as shown in the results from Zhou and Koltun~\cite{zhou2014color} in Figure~\ref{fig:teaser}.

\begin{figure}
    \centering
	\includegraphics[width=\linewidth]{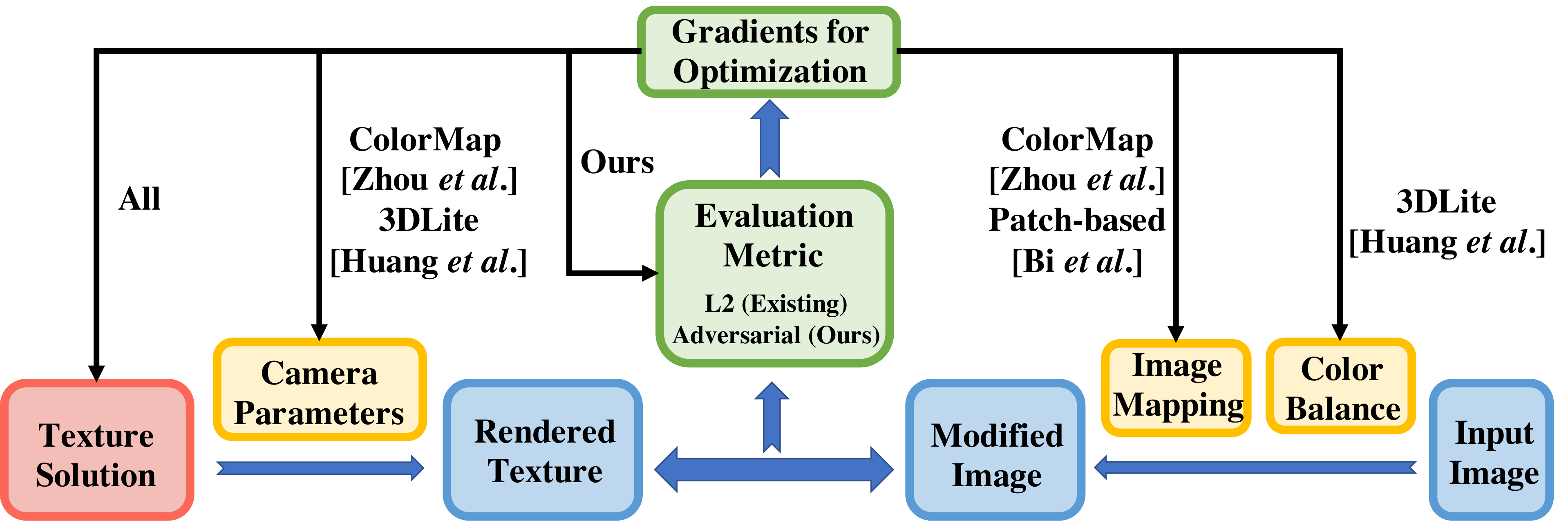}
	\caption{All methods target at optimizing a texture solution. Existing methods optimize the texture jointly with camera parameters~\cite{zhou2014color,huang20173dlite}, image mapping~\cite{zhou2014color,bi2017patch} or color balance~\cite{huang20173dlite}. Instead, we jointly solve texture with an adversarial evaluation metric to tolerate the errors.}
	\label{fig:concept}
\end{figure}

To address these issues, we propose a flexible texture optimization framework based on a learned metric that is robust to common scanning errors (right side of Figure~\ref{fig:teaser}).
 The key idea behind our approach is to account for misalignments in a {\em learned objective function} of the texture optimization.   
 Rather that using a traditional object function, like $L1$ or $L2$, we learn a new objective function (adversarial loss) that is robust to the types of misalignment present in the input data.  This novel approach eliminates the need for hand-crafted parametric models for fixing the camera parameters \cite{zhou2014color,huang20173dlite}, image mapping \cite{bi2017patch,zhou2014color}, or color balance \cite{huang20173dlite} (bottom row of Figure~\ref{fig:concept}) and replaces them all with a learned evaluation metric (green box in Figure~\ref{fig:concept}).   As such, it adapts to the input data.
 
Inspired by the success of adversarial networks in image synthesis~\cite{goodfellow2014generative}, we propose to use a learned conditional discriminator to serve our {\em objective function}, and jointly optimize the color texture of a reconstructed surface with this discriminator.
The condition is a captured image $I_A$ from the source view $V_A$, and the query is either (i) ``real:'' a second captured image $I_B$ (from an auxiliary view $V_B$) projected onto the surface and then rendered back to $V_A$, or (ii) ``fake:'' an image of the optimized synthetic texture rendered to view $V_A$. By optimizing the surface texture while jointly training this conditional discriminator, we aim to produce a texture that is indistinguishable from reprojections of captured images from all other views.  
During the optimization, the discriminator learns invariance to the misalignments and distortions present in the input dataset, while recognizing synthetic artifacts that do not appear in the real images (local blurs and seams).  Therefore, the textures optimized to fool the discriminator (ours in Figure~\ref{fig:teaser}) appear more realistic than in previous approaches.

Our experiments show that this adversarial optimization framework produces notably improved performance compared to state-of-the-art methods, both quantitatively on synthetic data and qualitatively on real data. 
Moreover, since it tolerates gross misalignments, we are able to generate realistic textures on CAD models which have been only roughly aligned to 3D scans, in spite of large mismatches in surface geometry. 
This opens up the potential to produce CAD models with realistic textures for content creation.

%% file: 1RelatedWorks.tex
\section{Related Work}
RGB-D scanning has a rich history, and many approaches have been proposed for color texture generation.

\vspace*{2mm}\noindent{\bf View aggregation.}
Common texture generation methods~\cite{izadi2011kinectfusion,zhou2014color} average projected input images to generate textures.  
To reduce blurriness artifacts, some approaches select a single or a few candidate views for each region \cite{dessein2014seamless}. Others formulate a multi-label selection energy minimization problem to minimize seam artifacts~\cite{lempitsky2007seamless,sinha2008interactive,velho2007projective,waechter2014let,huang20173dlite}. For instance, ~\cite{huang20173dlite} aims at selecting the best view for each region to balance the visual sharpness and color consistency of boundaries between neighboring regions with different views selected, which is modeled as a multi-label graph-cut problem~\cite{boykov2001fast}. Our method does not explicitly define the aggregation method, but implicitly aggregates colors from different views based on a learned adversarial metric.

\vspace*{2mm}\noindent{\bf Parametric color optimization.}
Several approaches have been proposed to improve the mapping of input images to textures with parametric models, 
leveraging both human supervision~\cite{franken2005minimizing,ofek1997multiresolution,pighin2006synthesizing,xu2019deep}, as well as automatic optimization~\cite{bernardini2001high,pulli2000surface}.  
Zhou \textit{et al.}~\cite{zhou2014color} propose to optimize a parametric model comprising camera poses and non-rigid grid deformations of input images to minimize an L2 color consistency metric. 
While these methods are able to fix small misalignments, their deformation models are often not expressive enough to handle many real-world distortions, particularly those due to largely approximate surface geometry. 
In contrast to a hand-crafted deformation model, we learn a distortion-tolerant adversarial loss.

\vspace*{2mm}\noindent{\bf Patch-based color optimization.}
Patch-based image synthesis strategies have been proposed for color texture optimization~\cite{bi2017patch}. Rather than non-rigid image warping, they re-synthesize the input image with the nearest patch~\cite{simakov2008summarizing} to handle misalignments. 
However, general misalignment cannot be accurately modeled by translating patches, and the L2 loss is not robust to color, lighting or sharpness differences. 
Our method optimizes the discriminator to cover all these problems without requiring explicit re-synthesis.

\vspace*{2mm}\noindent{\bf Neural textures.}
Recently, neural rendering approaches have been proposed to synthesize a feature map on a surface that can be interpreted by a deep network to produce novel image views.  For instance, \cite{thies2019deferred} stores appearance information as high-dimensional features in a neural texture map associated with the coarse geometry proxy and decodes to color when projected to novel views.  \cite{sitzmann2019deepvoxels} stores the appearance information as high-dimensional features in volumes, and \cite{aliev2019neural} uses features stored with points. These methods rely on the representation power of generative networks at rendering times to obtain novel viewpoints, which limits their applicability in standard graphics pipelines.

%% file: 2Approach.tex
\section{Method}

\begin{figure*}
    \centering
    \includegraphics[width=\linewidth,height=0.32\linewidth]{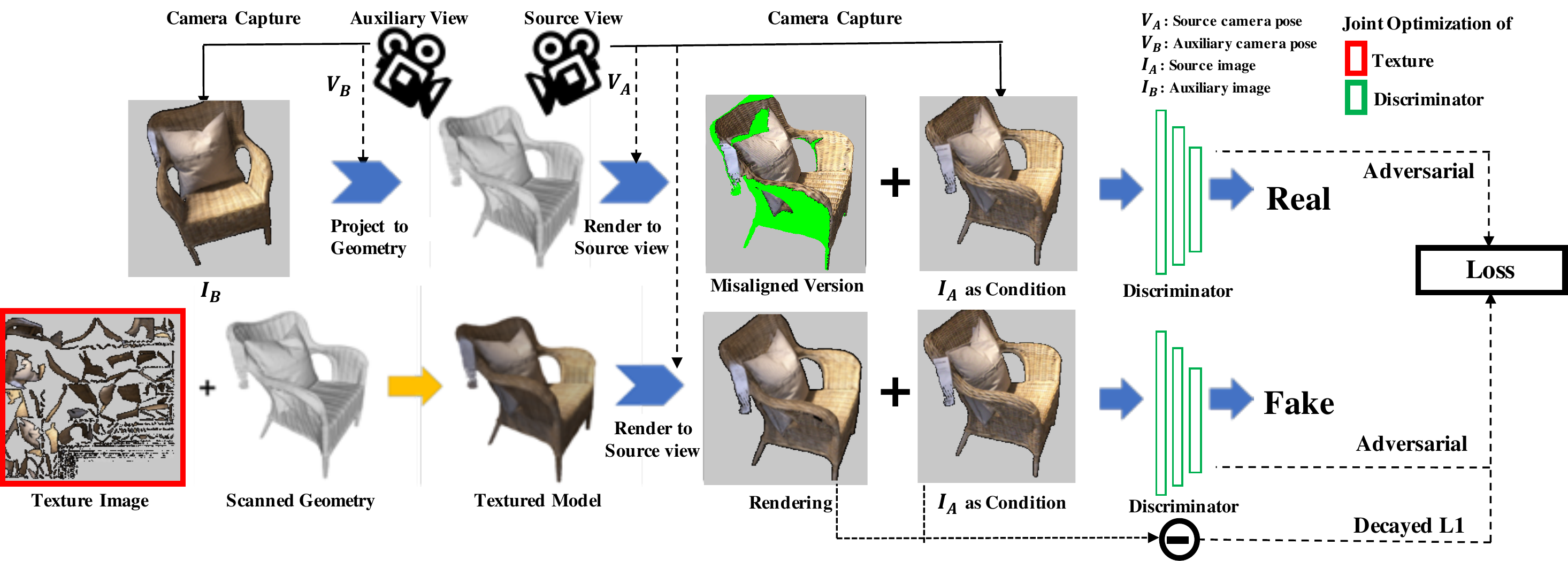}
    \caption{Texture Generation. 
    From an input RGB-D scan, we optimize for both its texture image and a learned texture objective function characterized by  a discriminator network. 
    The discriminator operates on reprojections of input color images in order to maintain robustness to various misalignments.
    We randomly pick a pair of input images, \emph{source} and \emph{auxiliary}, and synthesize the fake and real examples from the source view, conditioned on the re-projected source image. 
    The texture image and discriminator are trained in an alternating process.}
    \label{fig:pipeline}
\end{figure*}

Our goal is to optimize for a color texture that can be used to render a scanned scene using a classical computer graphics pipeline.
During the scanning procedure, we obtain color images and their estimated camera poses.
These views, along with the reconstructed geometry, are input to our method.
To optimize for a texture, we must specify an objective function; in this case, we must account for misalignments of the color images and the reconstructed model.
Thus we propose to learn the loss function in conjunction with the texture (see Figure~\ref{fig:pipeline}).
The function is modeled as an adversarial loss using a discriminator network to identify `real' and `fake' imagery, and is designed to provide a misalignment-tolerant metric for our texture optimization.

\subsection{Misalignment-Tolerant Metric}
\label{sec:approach-misalign}
\begin{figure}
    \centering
    \includegraphics[width=\linewidth]{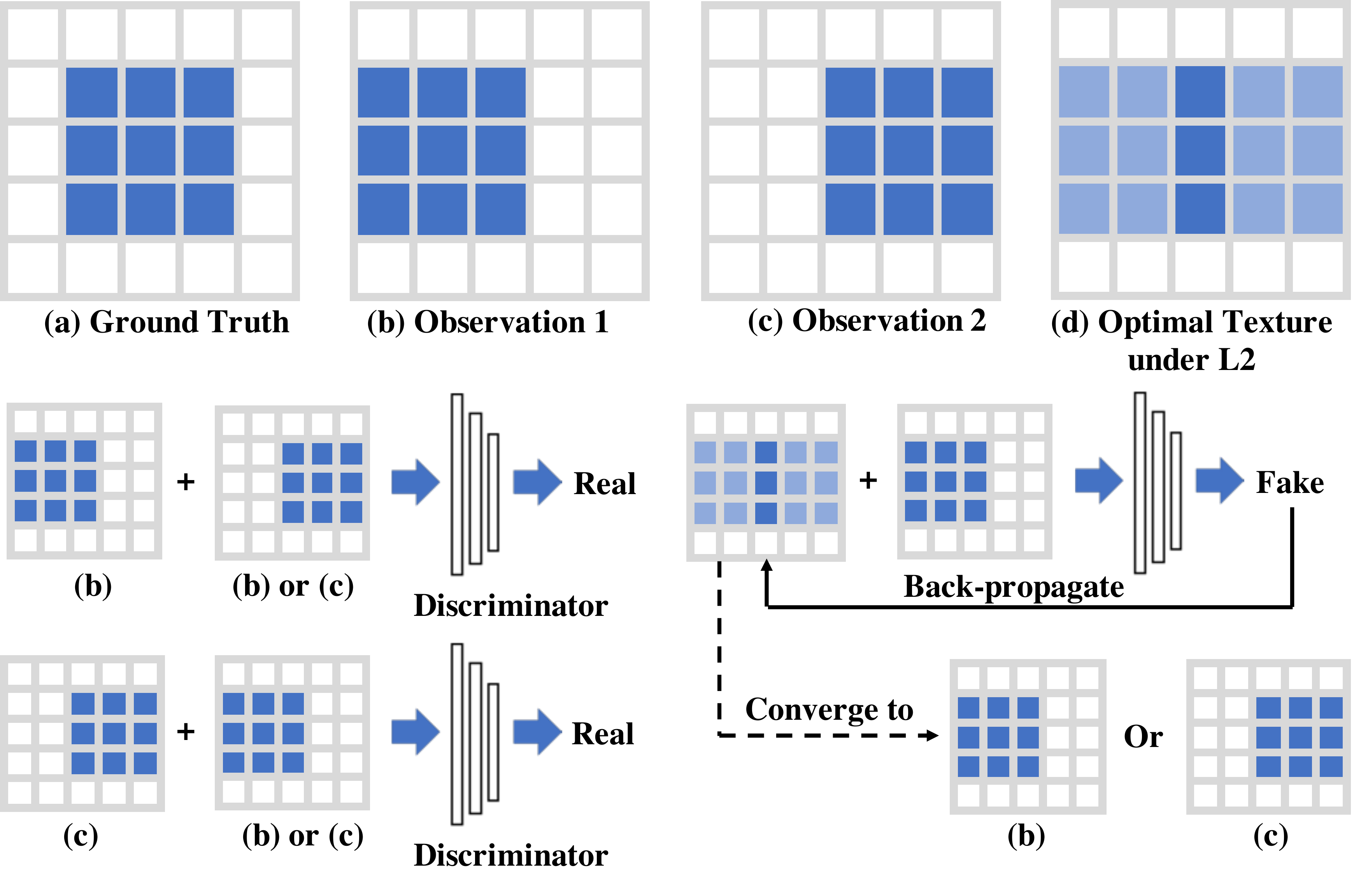}
    \caption{2D example of a misalignment. (a) shows the ground truth pattern, which is observed with misalignment in (b) and (c); an L2 loss results in blurring (d). 
    We train a discriminator which only accepts (b) and (c) as real examples conditioned on each other, and use it to optimize the texture, which converges to either (b) or (c).}
    \label{fig:misalign-example}
\end{figure}

Our key insight is to propose to learn a conditional discriminator as a misalignment-tolerant metric adaptive to the error distribution of the input data.
Figure~\ref{fig:misalign-example}(a) shows a 2D example where two observations (b) and (c) are misaligned by 2 units in the horizontal directions, and an L2 loss results in blurry appearance.
Ultimately, we aim to synthesize a texture that appears as realistic as either observation.
To achieve this, we ask the discriminator to consider both (b) and (c) as real conditioned on either observation.
With such a discriminator, the blurred (d) results in a large loss and the texture will instead converge to either (b) or (c).
We extend this intuition to 3D where the geometry is observed from different viewpoints.
We then aim to optimize a texture such that local patches of the texture rendered to various views look realistic.
Therefore, conditioned on any arbitrary view, we generate real examples by a re-projection from any other view to this view, as shown in Figure~\ref{fig:pipeline}.
Such re-projection can be achieved by projecting the color image onto the surface and then rendering back to another view.
Unlike the simple 2D example, it is highly possible that there is no texture solution so that each local patch perfectly matches the one view from the input images, given camera and geometry error.
However, the proposed approach is expected to push those inconsistencies to the smooth textured regions to hide any artifacts that can be easily identified by the discriminator, and thereby producing locally consistent realistic texture solution.
For each optimization iteration, we randomly select two input images, $I_A$ (source image) and $I_B$ (auxiliary image) with corresponding camera poses $V_A$ and $V_B$.
The conditioning is $I_A$ from the viewpoint $V_A$, and the `real' image is $I_B$ projected to the scan geometry and rendered from $V_A$, while the `fake' image is the synthesized texture rendered from $V_A$.
We alternating optimize the texture and discriminator.
During texture optimization, we adjust the texture pixel colors to maximize the adversarial loss such that it looks more realistic under the discriminator scoring.
During discriminator optimization, we minimize the adversarial loss such that it better classifies real and fake examples.
We linearly combine adversarial loss with an L1 loss that decays exponentially as the optimization proceeds, which helps the optimizer find a good initial texture solution.

\paragraph{Network Architecture}
Our framework is adopted from the PatchGAN discriminator architecture proposed by Isola et al.~\cite{isola2017image}.  We choose that framework because it is designed to produce local details that look as realistic as a given set of input images.
We use three convolutional layers, resulting in a patch size of $70\times 70$, which we find suitable for our input images of resolution $640\times 480$.
We apply a PatchGan to evaluate local $70\times 70$ patches of images rather than the entire image.
Patches are selected for discriminator training if more than half of the patch is not occluded. Thus, patches used for training have sufficient overlap.
Unlike the original, we remove all batch normalization layers and feed a single view example for each optimization iteration, which we empirically found to improve performance.
Conditioned on the input view, we ask the discriminator to evaluate the residual of the synthesized example subtracted by the condition input.
Finally, since we focus on evaluating foreground regions (pixels corresponding to input geometry), we remove the loss terms for regions where background comprises more than $90\%$ of the receptive field.

\subsection{Texture Optimization}

To retrieve a texture, we jointly optimize the texture and the misalignment-tolerant metric.
Inspired by the adversarial loss used in Pix2Pix~\cite{isola2017image}, we express our view-conditioned adversarial loss as:
\begin{align}
\begin{split}
    \mathcal{L}_c(T,D) &= \mathbb{E}_{x,y}(\log D(x,y)) +\\ &\mathbb{E}_{x,M_x}(\log (1 - D(x, M_x(T) ) ),
\end{split}
\end{align}
where $T$ and $D$ represent the target texture image and the discriminator parameters we are optimizing for.
$x$ is the condition, a reprojected color image from the input sequence of captured images.
$M_x$ is the fixed texture-to-image mapping given the camera pose associated with $x$. 
Here, a real example is an image $y$ re-projected to the view of $x$.
We optimize $D$ with the objective to correctly identify real examples, misaligned real imagery, and fake examples rendered from the texture as $M_x(T)$. 
Simultaneously, we optimize the texture $T$ such that it is difficult to be identified as fake when mapped to view of $x$.
Since the adversarial loss alone can be difficult to train, we additionally add an L1 loss to the texture optimization to provide initial guidance for the optimization:
\begin{equation}
\mathcal{L}_{L1}(T) = \mathbb{E}_{x,y,M_x} ||y - M_x(T)||_1.
\end{equation}
Our objective texture solution is:
\begin{equation}
    T^* = \arg \min_{T} \max_{D} \mathcal{L}_c(T,D) + \lambda \mathcal{L}_{L1}(T).
\end{equation}

During training, we initialize all pixels in texture image to zero and $\lambda=10$.
The high $\lambda$ allows the L1 loss to provide an initial texture, and for every 1000 steps we exponentially decay the lambda by a factor of $0.8$.
We optimize in alternating fashion for each optimization step, using the Adam optimizer for both the texture and discriminator with learning rates $10^{-3}$ and $10^{-4}$ respectively.
For each object or scene, we optimize for 50000 steps to finalize our texture. 

\subsection{Differentiable Rendering and Projection}
To enable the optimization of the RGB texture of a 3D model, we leverage a differentiable rendering to generate synthesized `fake' views.
We pre-compute a view-to-texture mapping using pyRender~\cite{huang2019framenet}, and can then implement the rendering with a differentiable bilinear sampling. 

To create the misaligned `real' images ($I_B$ seen from $V_A$), we compute a reprojection; note that here we do not need to maintain gradient information.
For each pixel $\mathbf{P}_A$ in the source image, we need to determine the corresponding pixel $\mathbf{P}_B$ in the auxiliary image, so that a bilinear sampling can be applied to warp image from the $V_B$ to $V_A$. 
Specifically, for $\mathbf{P}_A$ with depth value $d_A$ from the source depth map, we can determine its 3D location in the source view's space as $\mathbf{p_A}=d_A\mathbf{K}^{-1}\mathbf{P_A}$ where $\mathbf{K}$ is the intrinsic camera matrix. 
Suppose the transformations from the camera to the world space for the source and the auxiliary views are given as $\mathbf{T}_A$ and $\mathbf{T}_B$, the corresponding 3D and pixel location in the auxiliary view are $\mathbf{p}_B=\mathbf{T}_B^{-1}\mathbf{T}_A\mathbf{p_A}$ and $\mathbf{P}_B=\mathbf{K}\mathbf{p}_B$. 
The pixel is visible in the auxiliary view if $\mathbf{P}_B$ is in the scope of the image and the difference between z-dimension of $\mathbf{p}_B$ and $d_B$ from the auxiliary depth map is $<\theta_z$. We use $\theta_z=0.1$ meters for scenes $\theta_z=0.03$ for object level scanning.

%% file: 3Experiments.tex
\section{Experiments}

\paragraph*{Evaluation Metric}
For evaluation, we adopt several different metrics to measure the quality of the generated texture compared to the ground truth. 
First, we propose the nearest patch loss as an indicator of how close the patch appearance of the texture is to the ground truth. Specifically, for each pixel $\mathbf{u}$ we extract a $7\times 7$ patch centered around it in the generated texture and find the L2 distance $d(\mathbf{u})$ between it and the nearest neighbor patch in the ground truth texture. We define the nearest patch loss as the average of all $d(\mathbf{u})$.
Second, we adopt the perceptual metric~\cite{zhang2018unreasonable} to evaluate perceptual quality. 
Finally, we propose to measure the difference between generated textures and ground truth according to sharpness~\cite{vu2011bf} and the average intensity of image gradients, in order to evaluate how robust the generated textures are to blurring artifacts without introducing noise artifacts. 
Note that standard image quality metrics such as the mean square error, PSNR~\cite{de2003improved} or SSIM~\cite{brunet2011mathematical} are ill-suited, as they assume perfect alignment between target and the ground truth \cite{zhang2018unreasonable}.

\begin{figure}
    \centering
    \includegraphics[width=\linewidth]{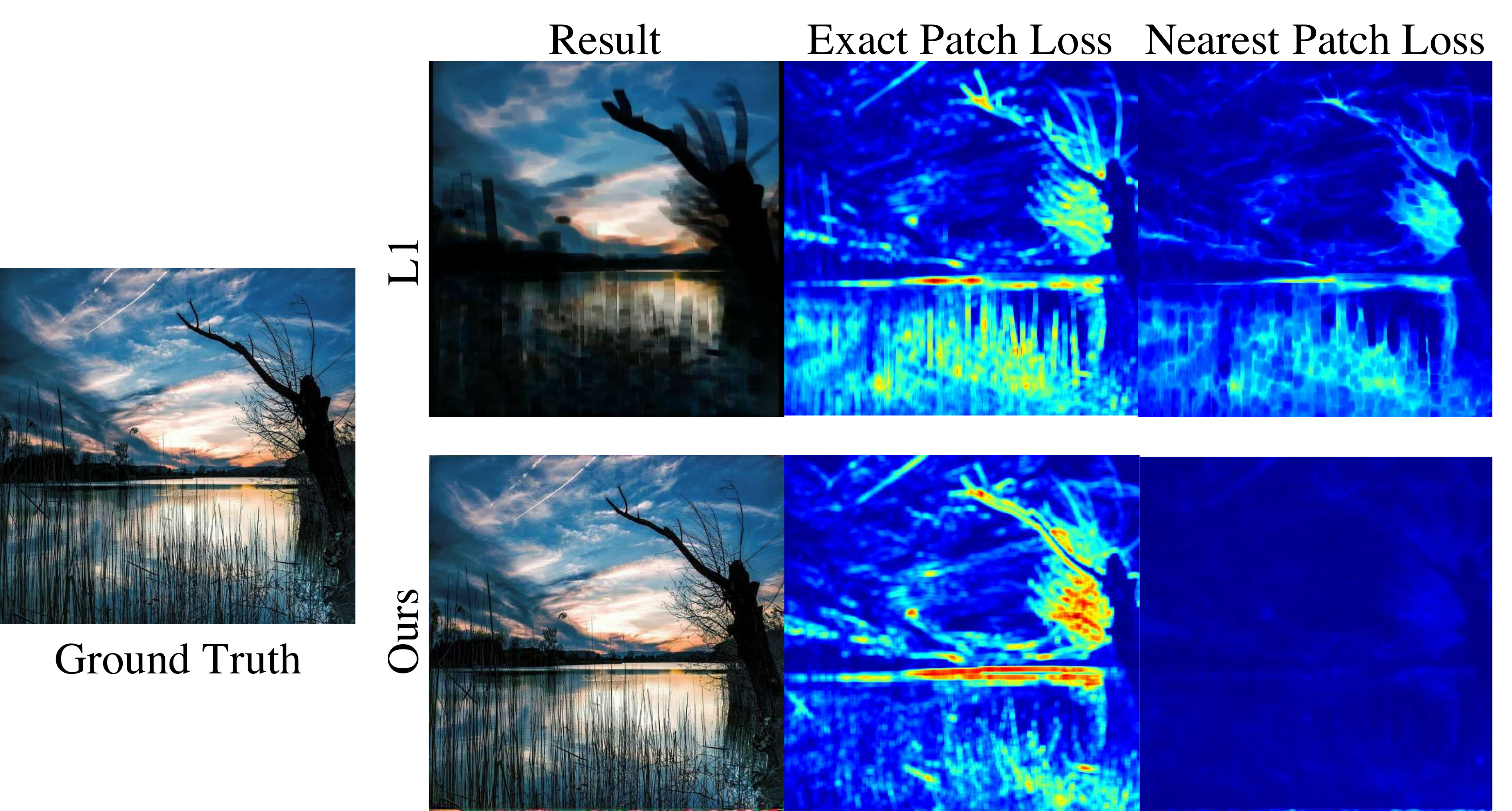}
    \caption{Texture Generation on 2D. The texture provided by our approach is visually closer to the ground truth image while avoiding blurring artifacts such as those introduced by an L1 loss.
    An exact patch loss favors alignment over perceptual similarity, while the nearest patch loss is a more robust metric.
    }
    \label{fig:2d-example}
\end{figure}

\paragraph*{Synthetic 2D Example}
We first verify the effectiveness of our method with a synthesized 2D example. 
We aim to optimize for a 2D image, given input observations with 2D micro-translation errors.
We use an image resolution of $512\times 512$ and translation error $\in[-16,16]^2$. 
During texture optimization,  we  randomly select one observation as the source and another observation as the auxiliary, and optimize the target image to be more realistic under the current discriminator. 
Figure~\ref{fig:2d-example} shows the resulting image optimized with our approach in comparison to a naive L1 loss.

Visually, our optimized image is sharper and perceptually closer to the ground truth while an  L1 loss results in blurry effect from aggregating multiple misaligned observations. 
In this simple setting, we evaluate the exact patch loss for each pixel quantitatively as the L2 distance of patches centered at this pixel between the generated image and the same one in the ground truth. The exact overall exact patch loss is the L2 norm of exact patch losses for all pixels. We additionally evaluate the nearest patch loss.
Optimization with the L1 loss achieves 10.7 exact patch loss while ours is 11.3. However, we achieve 1.53 nearest patch loss, which is smaller than L1 as 7.33. This suggests that our method prefers realistic misalignment to blur. We successfully derive an image where every local patch is nearly identical to a misaligned version of the patch in the ground truth image.

\paragraph*{Synthetic 3D Example}
In order to quantitatively evaluate our 3D texture generation, we create a synthetic dataset of 16 models randomly selected from ShapeNet~\cite{chang2015shapenet} across different categories. 
These shapes typically contain sharp edges and self-occlusion boundaries, complexities reflecting those of real-world objects.
Since we aim to address arbitrary texturing, we enrich the appearance of these shapes by using 16 random color images from the internet as texture images. 
To create  virtual scans of the objects, we uniformly sample $>900$ views on a unit hemisphere by subdividing an icosahedron, from which we render the textured geometry as observed color images. 
To simulate misalignment, we associate each rendered image with a slightly perturbed camera pose, and to simulate geometry errors, we apply random perturbations to the geometric model.
We use a set of errors increasing from $n=1$ to $n=4.5$, and refer to the supplemental material for additional detail regarding generating camera and geometry perturbations.

\begin{figure}
    \centering
    \includegraphics[width=\linewidth]{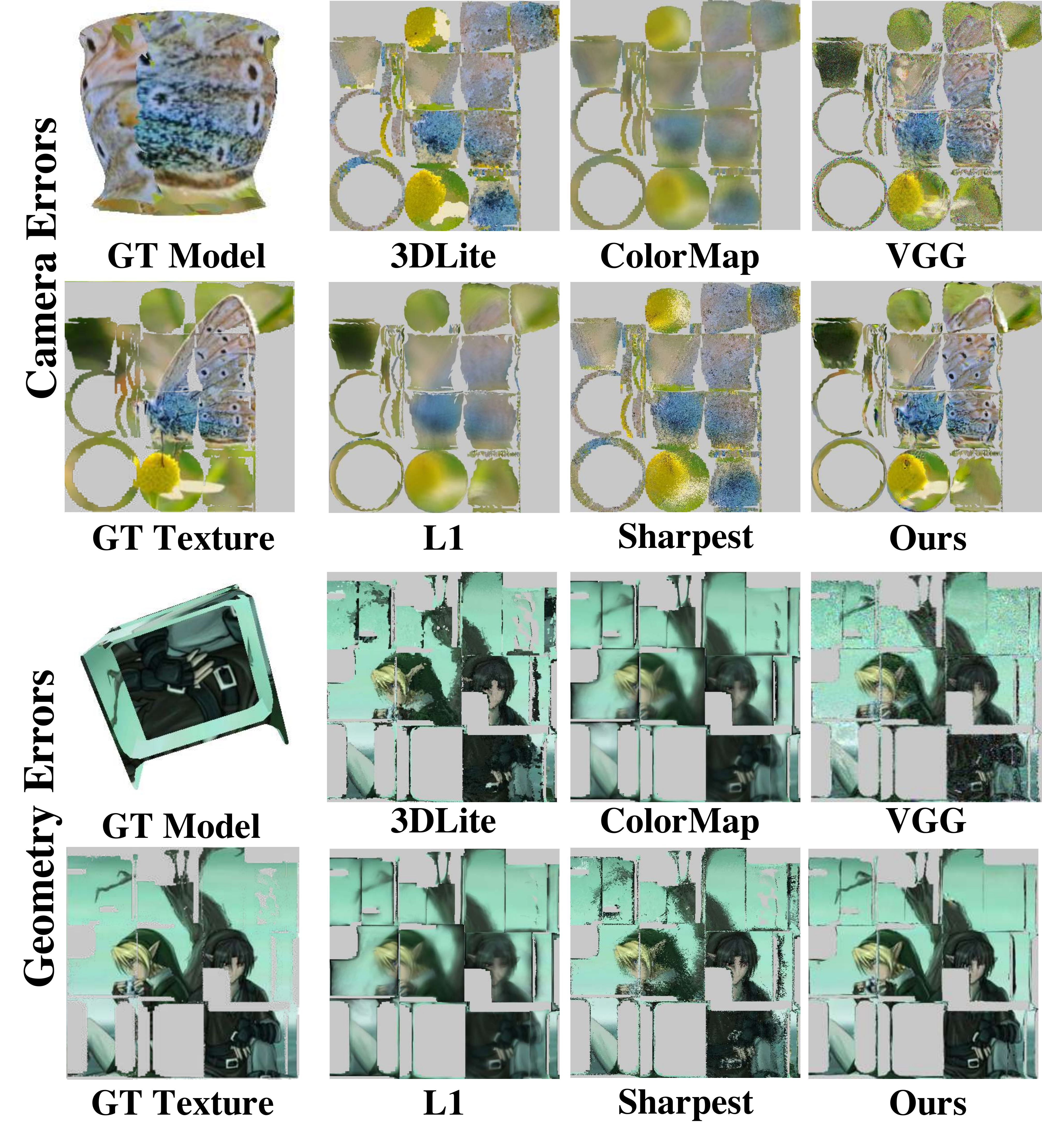}
    \caption{Texture generation in case of high camera or geometry errors. ColorMap~\cite{zhou2014color} suffers from blurring, and Sharpest or 3DLite~\cite{huang20173dlite} selection lead to inconsistent boundaries or breaks structures. VGG~\cite{johnson2016perceptual} aggregates views by blending deep features with noises, which is not sufficient for handling misalignment artifacts. Ours is visually closest to the ground truth.}
    \label{fig:pose-visual}
\end{figure}

In Table~\ref{tab:cam-err}, we study the effect of varying camera and geometry errors in this synthetic 3D setting.  We report evaluation metrics for our approach as well as several state-of-the-art texture optimization methods, including methods based on an L1 loss and texturing using sharpest frame selection~\cite{vu2011bf}.
Our approach outperforms all other methods, as it avoids blurring effects often seen with L1 and ColorMap~\cite{zhou2014color}, and it avoids seams and over-sharpness introduced by methods relying on sharpness selection (3DLite~\cite{huang20173dlite} and sharpest frame selection).
VGG~\cite{johnson2016perceptual} aggregates views by blending deep features, which is insufficient for handling misalignment artifacts.
Two example scenes with increasing errors in camera and geometry are shown in Figure~\ref{fig:pose-visual}.

\begin{table}
    \centering
    \includegraphics[width=\linewidth]{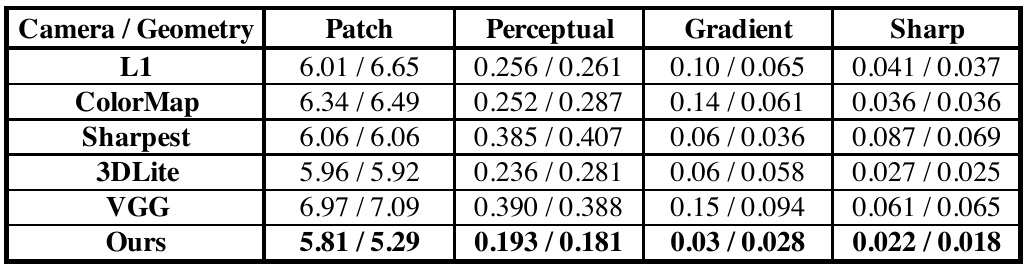}
    \caption{Evaluation of different methods on our 3D synthetic dataset averaged across different levels of camera pose and geometry errors.}
    \label{tab:cam-err}
    \vspace{-0.1in}
\end{table}

\begin{figure}
    \centering
    \includegraphics[width=\linewidth]{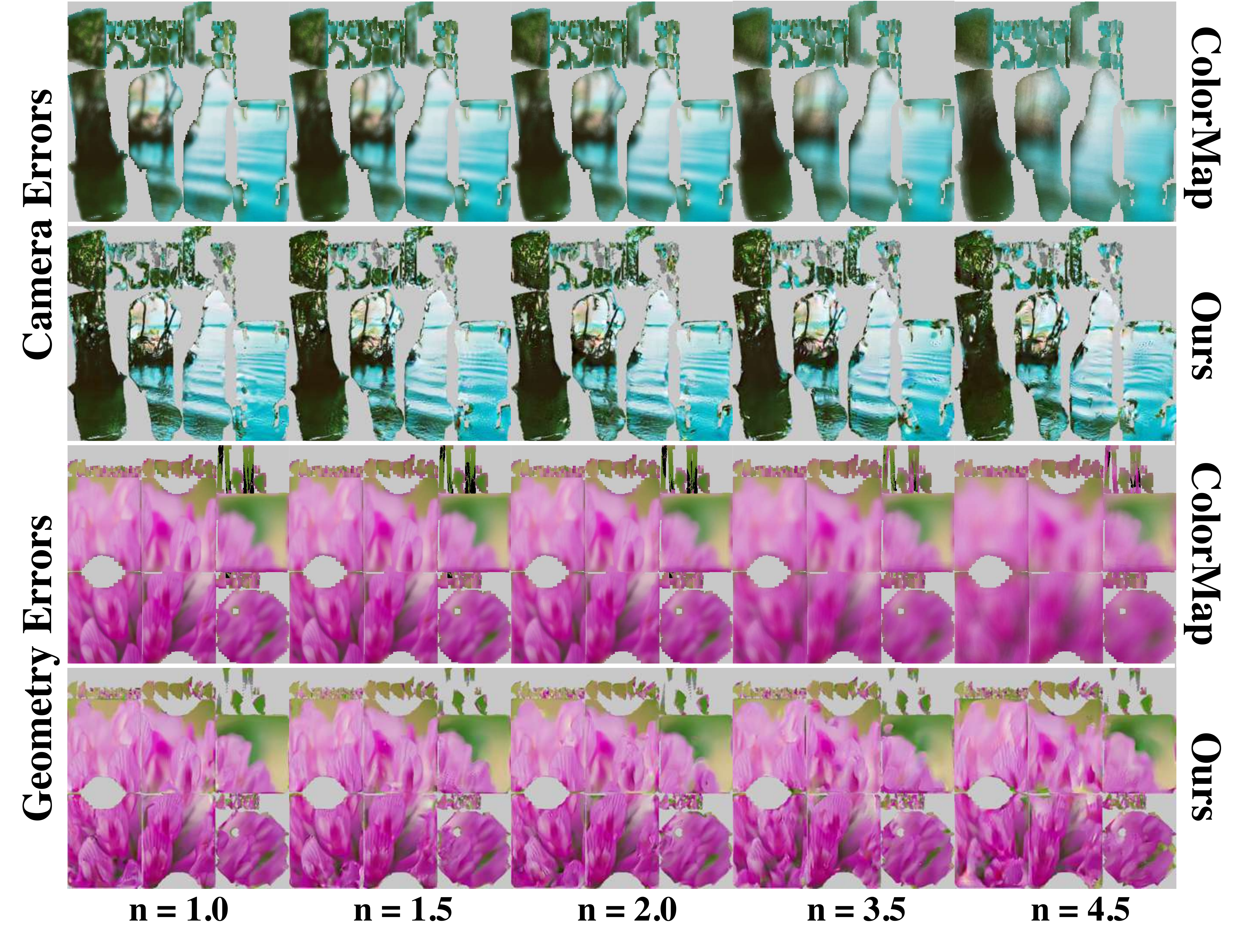}
    \caption{Texture generation under increasing camera or geometry errors. ColorMap~\cite{zhou2014color} produces more blurry results under camera/geometry error while ours maintains sharp textures.}
    \label{fig:pose-diff}
\end{figure}

\begin{figure}
    \centering
    \includegraphics[width=0.9\linewidth]{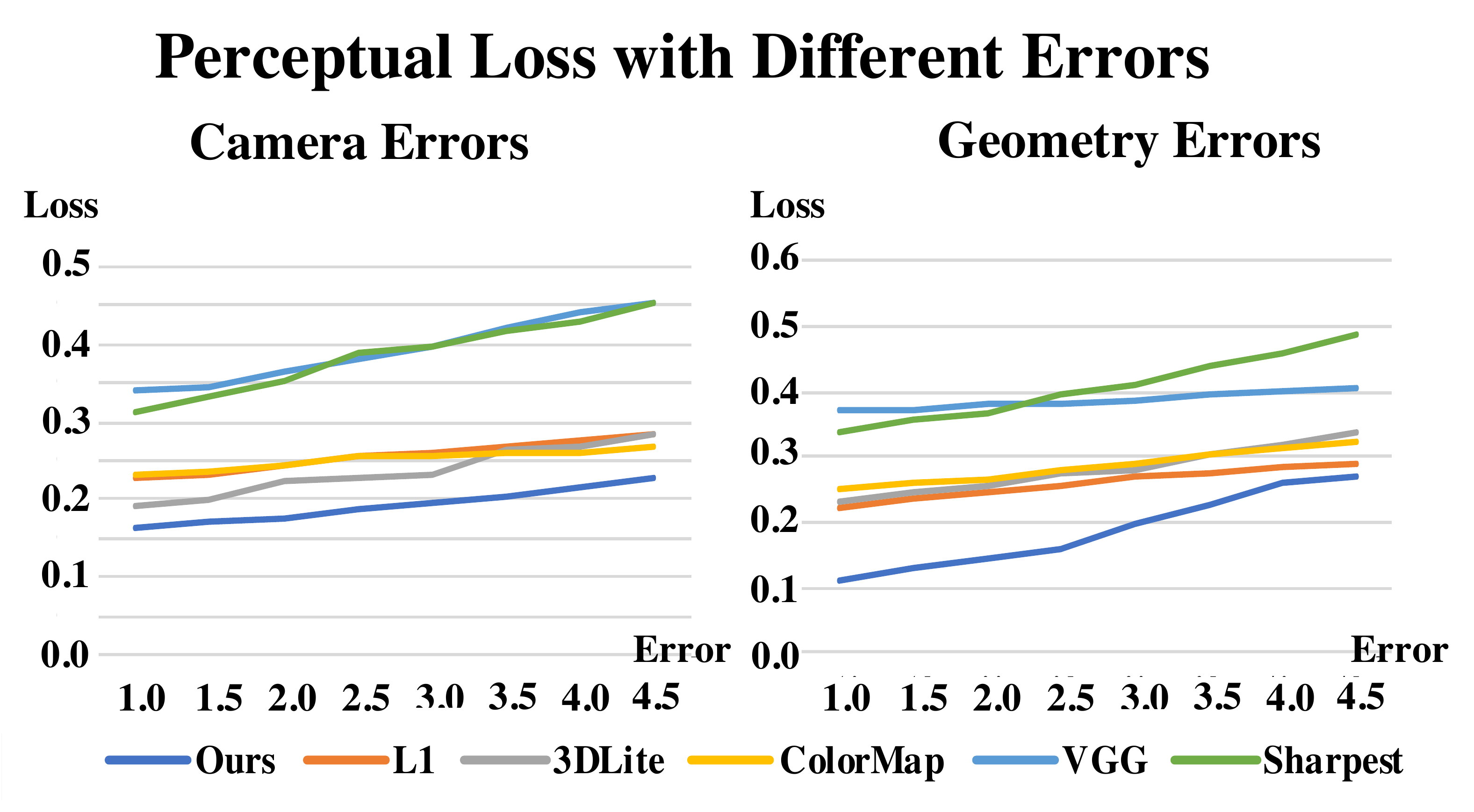}
    \caption{Perceptual loss under increasing camera or geometry errors; we outperform existing methods at various levels of error.}
    \label{fig:pose-chart}
\end{figure}

We additionally study the behavior of all methods in this experiment using the perceptual metric~\cite{zhang2018unreasonable} in Figure~\ref{fig:pose-chart}.
Although the performance drops for all methods with the increase of camera/geometry errors, our approach maintains the best perceptual quality as the errors increase. 
Figure~\ref{fig:pose-diff} shows a qualitative comparison; our approach maintains a sharp result while ColorMap produces increasingly blurriness as the error increases. 


\begin{figure}
    \centering
    \includegraphics[width=\linewidth]{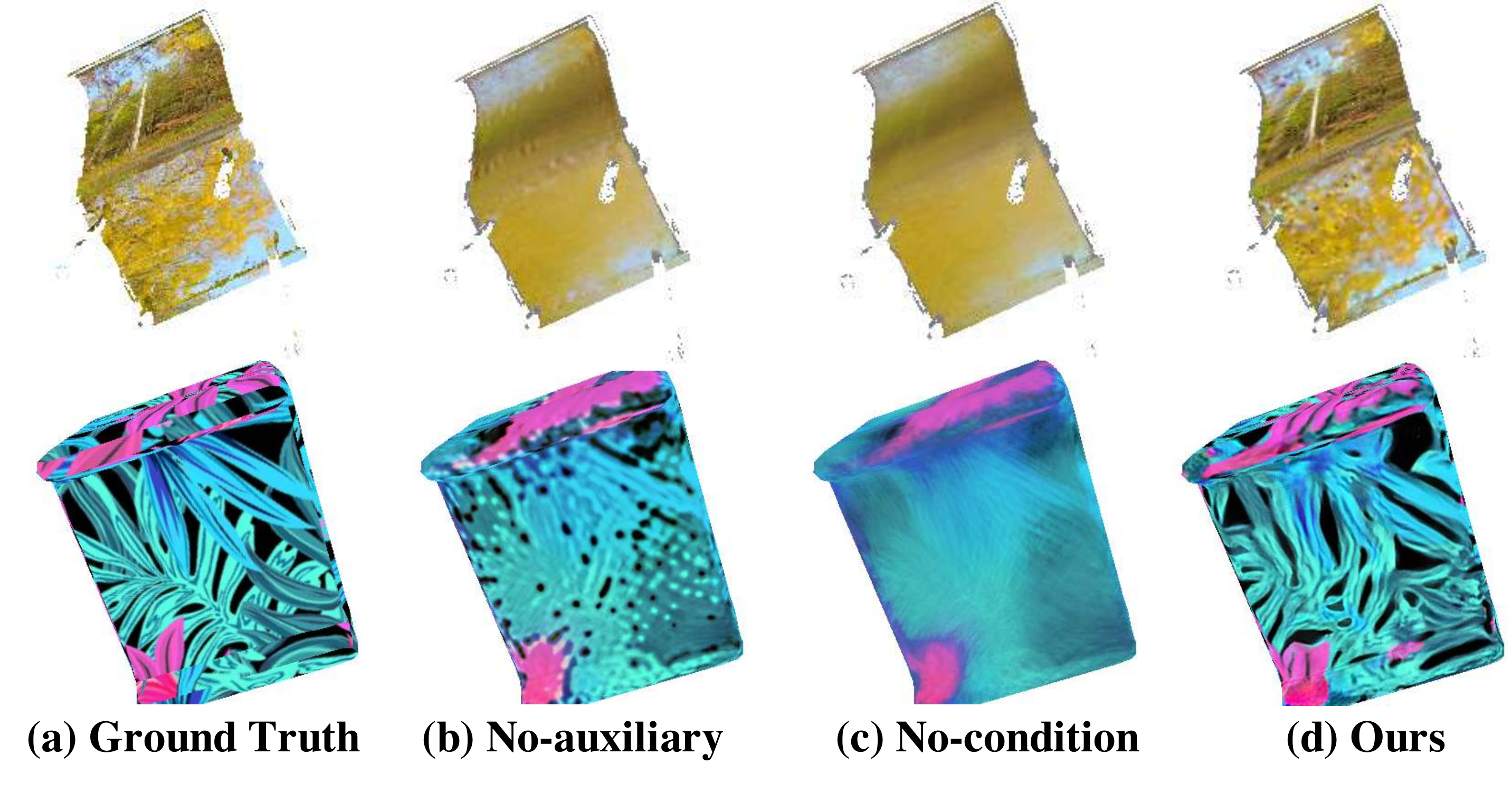}
    \caption{Comparing different discriminator options. (b) removes the auxiliary view from the discriminator, resulting in the lack of robustness to misalignments. (c) removes the condition from the discriminator, resulting in ambiguity in local regions. (d) our conditional discriminator leveraging auxiliary views to provide examples of realistic misalignments enables tolerance to misalignment and generation of textures reflecting input image characteristics. }
    \label{fig:gan-err}
\end{figure}

\paragraph*{Alternative Discriminators?} 
We analyze the design choices for our misalignment-tolerant conditional discriminator in Figure~\ref{fig:gan-err}.
Removing the auxiliary view (b) and thus relying only on the source view to provide `real' examples to the discriminator (similar to pix2pix~\cite{isola2017image}) renders the metric unable to handle misalignments.
We also evaluate a  general discriminator that classifies whether a generated patch is real or fake among entire input view sets without any condition (c), resulting in ambiguity as to where real patches come from. 
Our conditional discriminator leveraging reprojected auxiliary views enables robustness to misalignment, resulting in realistic texturing.

%% file: 4Result.tex
\begin{figure*}
    \centering
    \includegraphics[width=\linewidth]{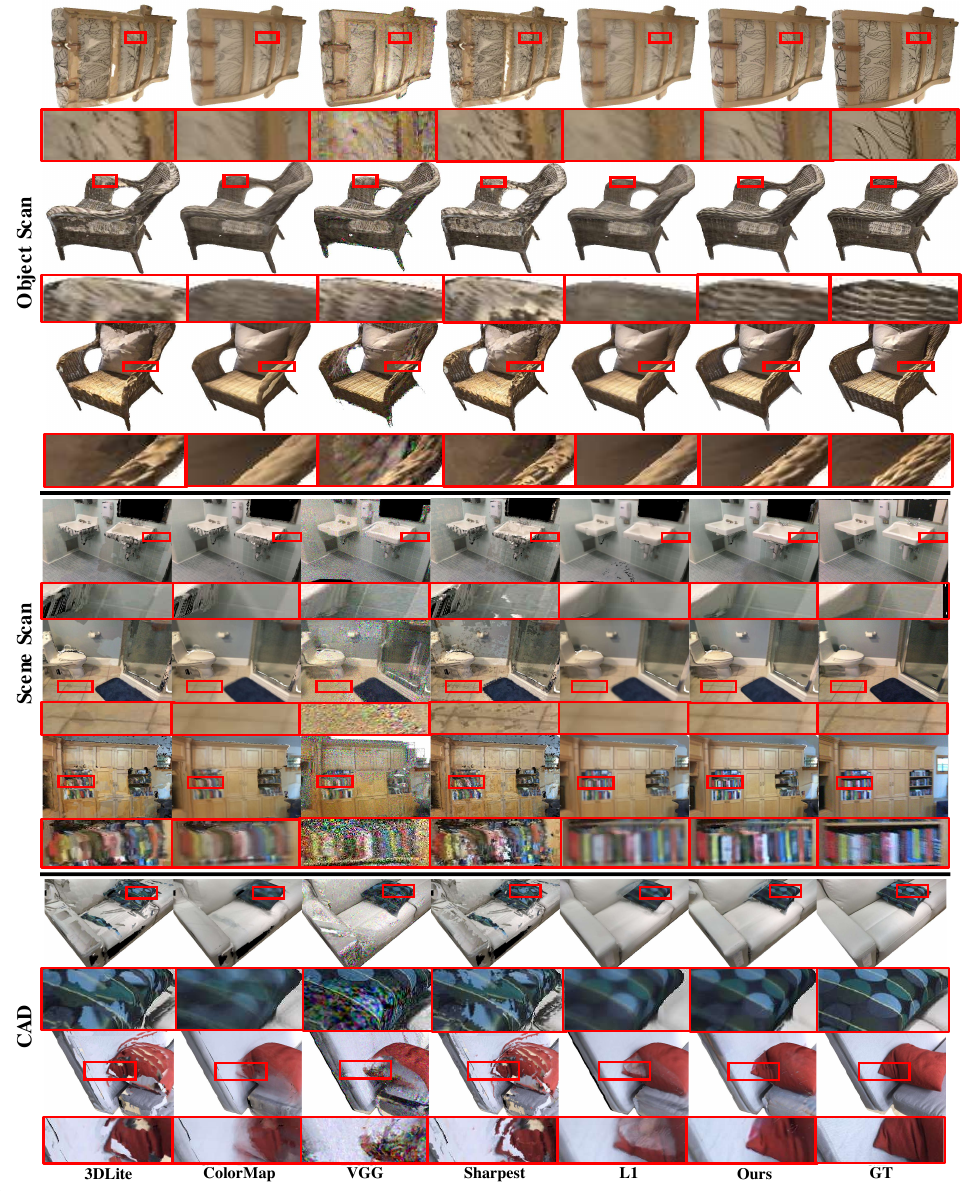}
    \caption{Visual comparison on object scans, ScanNet~\cite{dai2017scannet} scans of scenes, and CAD models aligned with object scans. Due to misalignment errors in both camera pose and geometry, both L1 loss and ColorMap~\cite{zhou2014color} produce blurry artifacts, sharpest selection and 3DLite~\cite{huang20173dlite} result in inconsistent regions or breaks in texture structure, and VGG~\cite{johnson2016perceptual} blends learned features resulting in structural artifacts and noise. Our misalignment-tolerant approach produces sharp and consistent textures.}
    \label{fig:object-scan}
\end{figure*}

\vspace{-0.5cm}
\paragraph*{Real Object Scans}
We compare our method to state-of-the-art texturing methods on scanned objects from real environments. We use a structure sensor~\cite{structuresensor} along with its SLAM system to scan 35 chairs, producing scanned geometry, RGB-D frames and the camera poses ($\approx 500$ frames per scan). 
The foreground/background for the object in the RGB frames is determined by whether a ray intersects with the reconstructed geometry. Figure~\ref{fig:object-scan} (rows 1-4) shows qualitative comparisons.
With an L1 loss or ColorMap~\cite{zhou2014color}, blur artifacts are induced by misalignment errors.
Sharpest selection and 3DLite~\cite{huang20173dlite} use sharp region selection, resulting in seams and inconsistent global structures, as shown in the flower, leaf, and chair arms. 
A VGG loss~\cite{johnson2016perceptual} produces excess noise artifacts.  
Our approach produces sharp and consistent texturing, including detailed patterns such as the leaves in row 1 and woven structures in rows 2 and 3.

Additionally, we show a quantitative evaluation in Table~\ref{tab:result} (first column) by evaluating the perceptual metric~\cite{zhang2018unreasonable} for rendered textures against input observed views; our approach achieves the most realistic texturing.

\paragraph*{Real Scene Scans}
To demonstrate the capability of our approach to optimize texture on a larger scale, we run our algorithm on the ScanNet dataset~\cite{dai2017scannet}, which provides RGB-D sequences and reconstructed geometry of indoor scenes.
We evaluate our approach on scenes with ID $\leq 20$ ($\approx 2000-3000$ frames per scan) and compare it with the existing state of the arts. 
Figure~\ref{fig:object-scan} (rows 5-9) and Table~\ref{tab:result} (middle column) show qualitative and quantitative comparisons. 
Our method produces texturing most perceptually similar to the observed images; our misalignment-tolerant metrics aids in avoiding blur, increased sharpness, or excess noise produces by other methods due to camera and geometry errors in real-world scans.

\paragraph*{Real to CAD Models}
Since our method can better handle errors from approximate surface geometry, it is possible to consider texturing CAD models using real-world images to attain realistic appearances.   While large datasets of 3D CAD models are now available~\cite{chang2015shapenet}, they are often untextured or textured simplistically, resulting in notably different appearance from real-world objects. 
To test whether our method can be applied in this challenging scenario, 
we use our collected dataset of real object scans, retrieve similar CAD models from ShapeNet manifold~\cite{huang2018robust}, and rigidly align them to the scanned objects.
We then replace the scanned geometry with the CAD model and then use the captured color images and estimated poses from the scan to optimize the CAD texture.
Qualitative and quantitative evaluation of our approach in comparison to existing state-of-the-art methods are show in Figure~\ref{fig:object-scan} (rows 10-13) and Table~\ref{tab:result} (right column), respectively.
Our approach is able to handle both camera poses errors as well as the synthetic-real geometry differences to produce texturing perceptually very similar to observed imagery, whereas other methods suffer strong blur, noise, and seam artifacts under these errors.

\paragraph*{Perceptual Quality}
\begin{table}
    \centering
    \footnotesize
    \begin{tabular}{|c|c|c|c|}
        \hline
        & Object & ScanNet & CAD\\
        \hline
        L1 & 0.197 & 0.470 & 0.199 \\
        \hline
        ColorMap & 0.186 & 0.461 & 0.234 \\
        \hline
        Sharpest & 0.222 & 0.510 & 0.260 \\
        \hline
        3DLite & 0.185 & 0.445 & 0.238 \\
        \hline
        VGG & 0.272 & 0.534 & 0.289 \\
        \hline
        Ours & \textbf{0.175} & \textbf{0.395} & \textbf{0.176} \\
        \hline
    \end{tabular}
    \vspace{0.1cm}
    \caption{Mean perceptual loss comparing the input images and rendered textures from different methods. Our method achieves best performance in the real and CAD datasets.}
    \label{tab:result}
\end{table}
\begin{figure}
    \centering
    \includegraphics[width=0.8\linewidth]{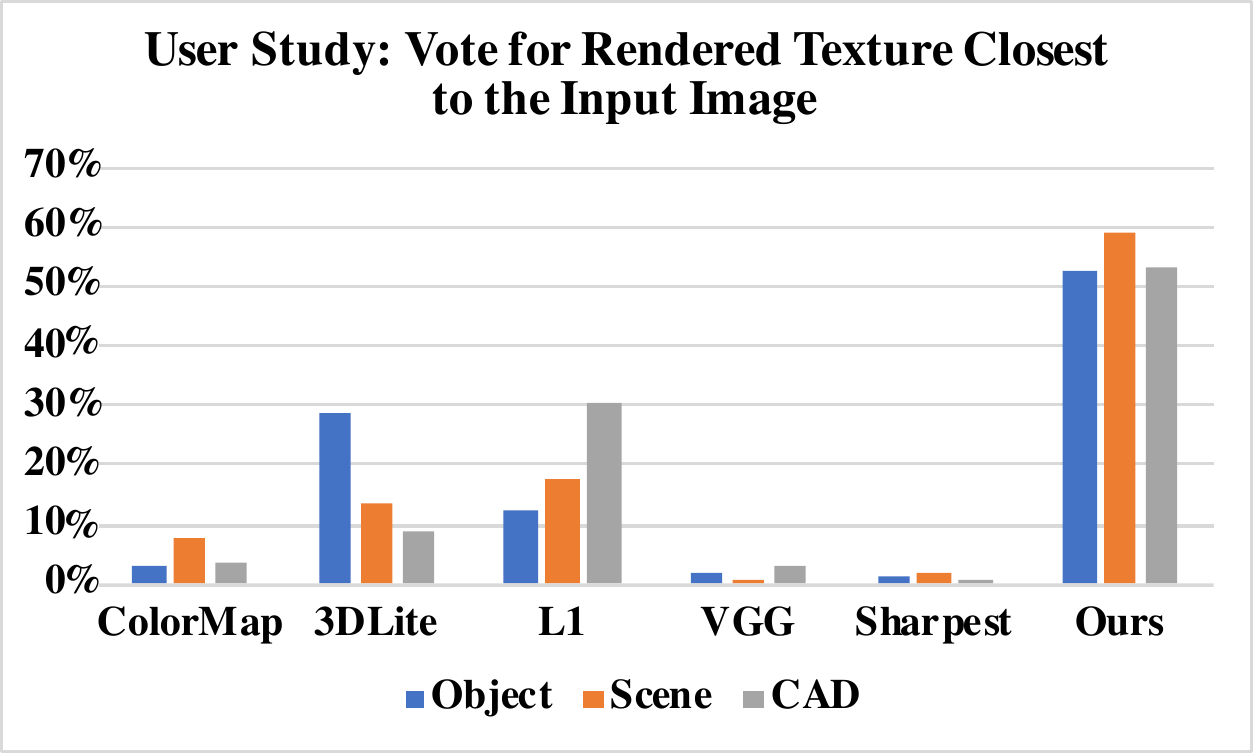}
    \caption{User study. We ask people to vote for the rendered textures from different methods that look closest to the input image.}
    \label{fig:user-study}
    \vspace{-0.1in}
\end{figure}
Although we lack ground truth texturing for objects in real environments, we can compare the perceptual loss~\cite{zhang2018unreasonable} of the rendered textured geometry from the corresponding viewpoint. We select 10 views uniformly distributed from the scanning video and render the textured model to compute the mean of the perceptual loss.
Table~\ref{tab:result} shows the performance of different methods on the object scans, scene scans and the CAD models; our method achieves the best performance in these three scenarios.

Additionally, we perform a user study to evaluate the quality of the texture, shown in Figure~\ref{fig:user-study}.
Our user study comprised $63$ participants who were asked to vote for the texture which produced a rendering closest to the input image. 
For some views, it can sometimes be difficult for users to differentiate between different methods when regions are largely uniform in color. 
Nevertheless, our method is  still notably preferred over other texturing approaches. We provide additional comparisons with ~\cite{waechter2014let} and ~\cite{fu2018texture} in supplemental C and describe the influence of sparse views for training discriminators in supplemental D.

\noindent \textbf{Runtime.}
On average, our released implementation takes 7.3 minutes per object and 33.4 minutes per scene on a single TITAN X GPU.

%% file: 5Conclusion.tex
\section{Conclusion}
We have proposed a misalignment-tolerant metric for texture optimization of RGB-D scans, introducing a learned texturing objective function for maintaining robustness to misalignment errors in camera poses and geometry.
We represent the learned function as a conditional discriminator trained with an adversarial loss where `real' examples characterize various misalignment errors seen in the input data.
This avoids explicit parametric modeling of scanning errors, and enables our optimization to produce texturing reflective of the realism.
Our approach opens up the potential for texturing synthetic CAD models with real-world imagery.
It also makes an important step towards creating digital content from real-world scans, towards democratized use, for instance in the context of AR and VR applications.

%% file: 6supplemental.tex
\clearpage
\section*{Supplemental}
\begin{appendix}
\section{Implementation details}
\subsection{Data Loader During Optimization}
Generating real and fake examples are two critical steps in our optimization process.

\begin{figure}[H]
    \centering
    \includegraphics[width=0.8\linewidth]{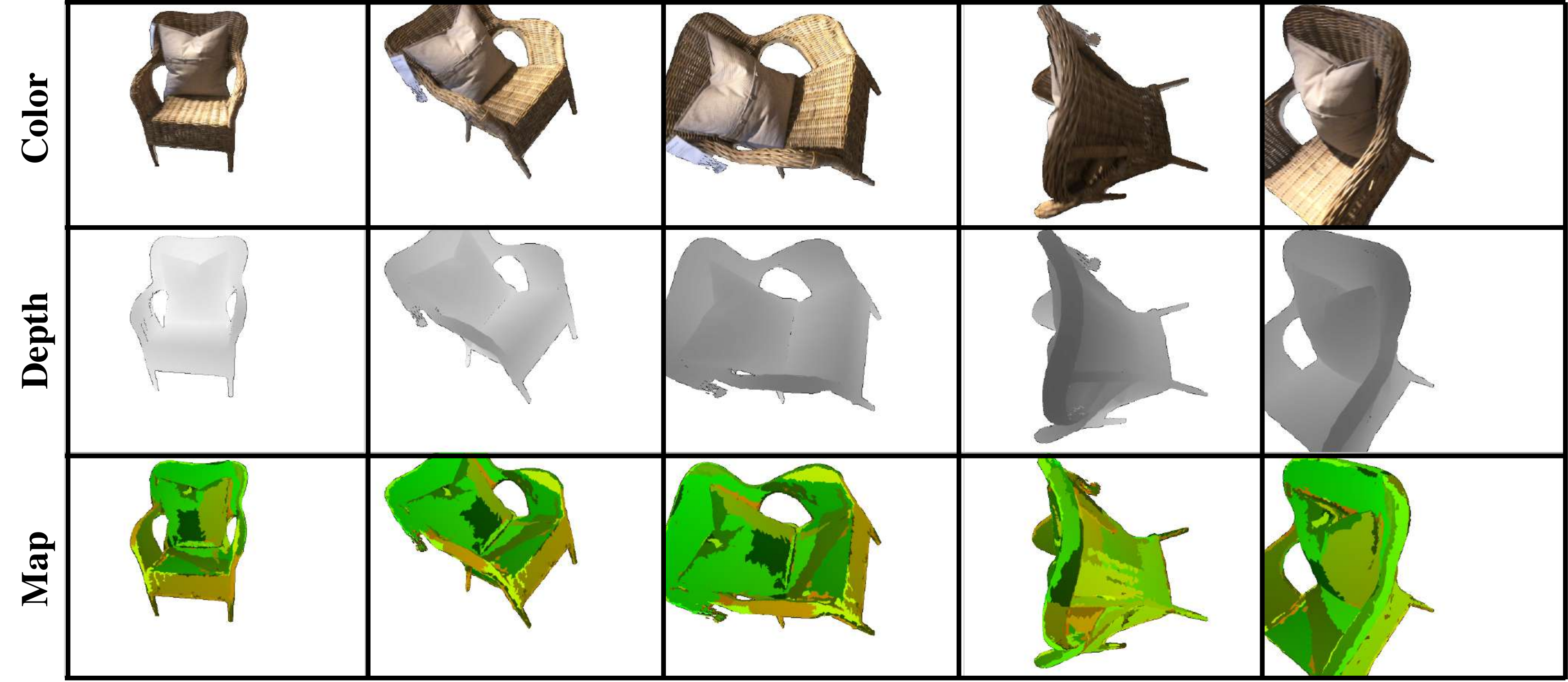}
    \caption{For each view of the scan, we cache the corresponding color and depth images. In settings where depth data is unavailable, we render depth maps from the target geometry. Additionally, we pre-compute the image-to-texture map for differentiable rendering. }
    \label{fig:dataset}
\end{figure}

To achieve this, we provide three images for each input view as shown in Figure~\ref{fig:dataset}. For each view, we provide the color and the depth image from the device, as shown in the first two rows. The background pixels can be removed simply by deciding whether the corresponding rays are intersection the reconstructed mesh. Additionally, we provide the view-to-texture mapping for the differentiable rendering of the texture image.

\begin{figure*}
    \centering
    \includegraphics[width=\linewidth]{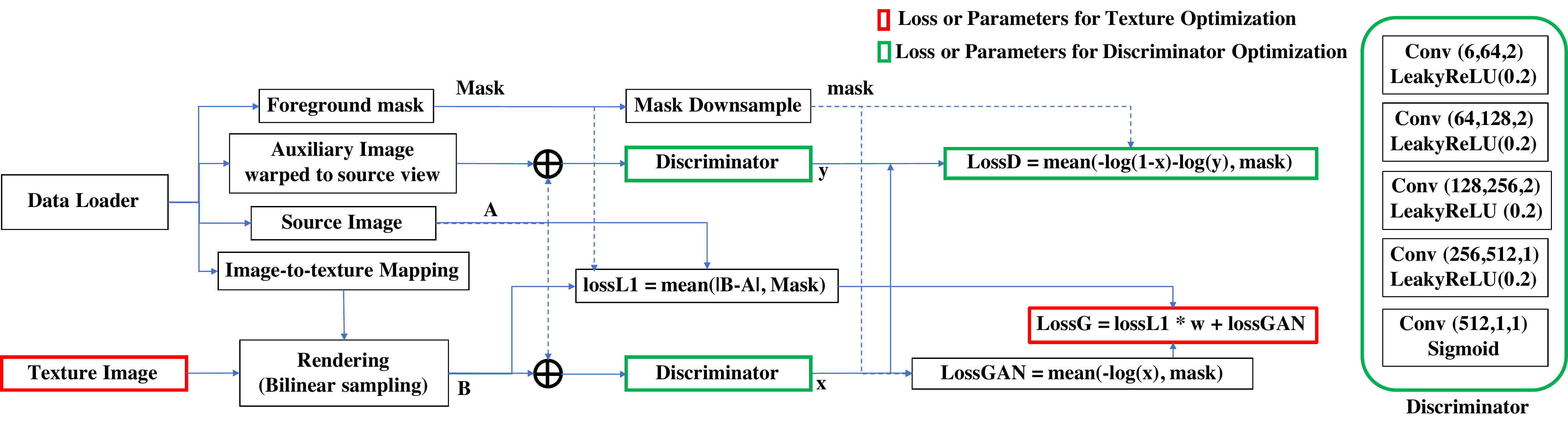}
    \caption{Detailed description of the optimization process.}
    \label{fig:details}
\end{figure*}
Figure~\ref{fig:details} shows the details of our optimization process. For each iteration, the data loader randomly selects a pair of related views as the source and the auxiliary, and we feed the network with a foreground mask containing the 3D scan, the real example synthesized by reprojecting the auxiliary image to the source view, the source image and the image-to-texture mapping. In the texture optimization stage, we render the target texture to source view based on image-to-mapping with a differentiable bilinear sampling as B. It is combined with the source image A as condition, sent to the discriminator to derive the prediction $x$. $x$ is used to compute the adversarial loss as ``lossGAN''. We additionally compute the L1 loss between A and B as ``L1'', and linearly combine ``lossGAN'' and ``L1'' with an exponentially decayed $w$ as ``lossG'', which is the objective function for optimizing the texture image. For every 1000 steps, we exponentially decay the $w$ by a factor of $0.8$. In the discriminator optimization stage, the real example is combined with source image A and sent to discriminator to derive the prediction $y$. $x$ from the fake example and $y$ from the real example are combined to compute the object adversarial loss ``lossD'' for discriminator optimization.

The discriminator architecture consists of 5 convolutional blocks (figure~\ref{fig:details}). Conv(x,y,z) represents a 4x4 convolution with padding as 1, input channel as $x$, output channel as $y$ and stride as $z$. Each convolution block is followed with a gate function where the first four are leakyReLUs and the last is sigmoid. 

\subsection{Details for Synthetic 3D Data Generation}
 We studied the behavior of our approach given the inaccurate camera pose or geometry. Camera perturbation is achieved by adding uniformly distributed noises to each dimension of the translation ranging from $[-e_t,e_t]$ and rotation as euler angle ranging from $[-e_a,e_a]$. To simulate geometry errors, we randomly generate a scalar for each vertex following the uniform distribution ranging from $[-e_g,e_g]$.
 Then, we apply 3 steps of Laplacian smooth to the scalars. We move the vertices along their normal directions with the distance specified by these scalars. We compare our method with different approaches for all selected ShapeNet objects with different amount errors. For camera errors, we set $e_t=0.01*1.5^n$ ($n\in\{1.5,2,2.5,3,3.5,4,4.5\}$) and $e_a=5^{\circ}$. For geometry errors, we set $e_g=0.02*1.5^n$ with $n\in\{1.5,2,2.5,3,3.5,4,4.5\}$.
 
\begin{figure*}
    \centering
    \includegraphics[width=0.9\linewidth]{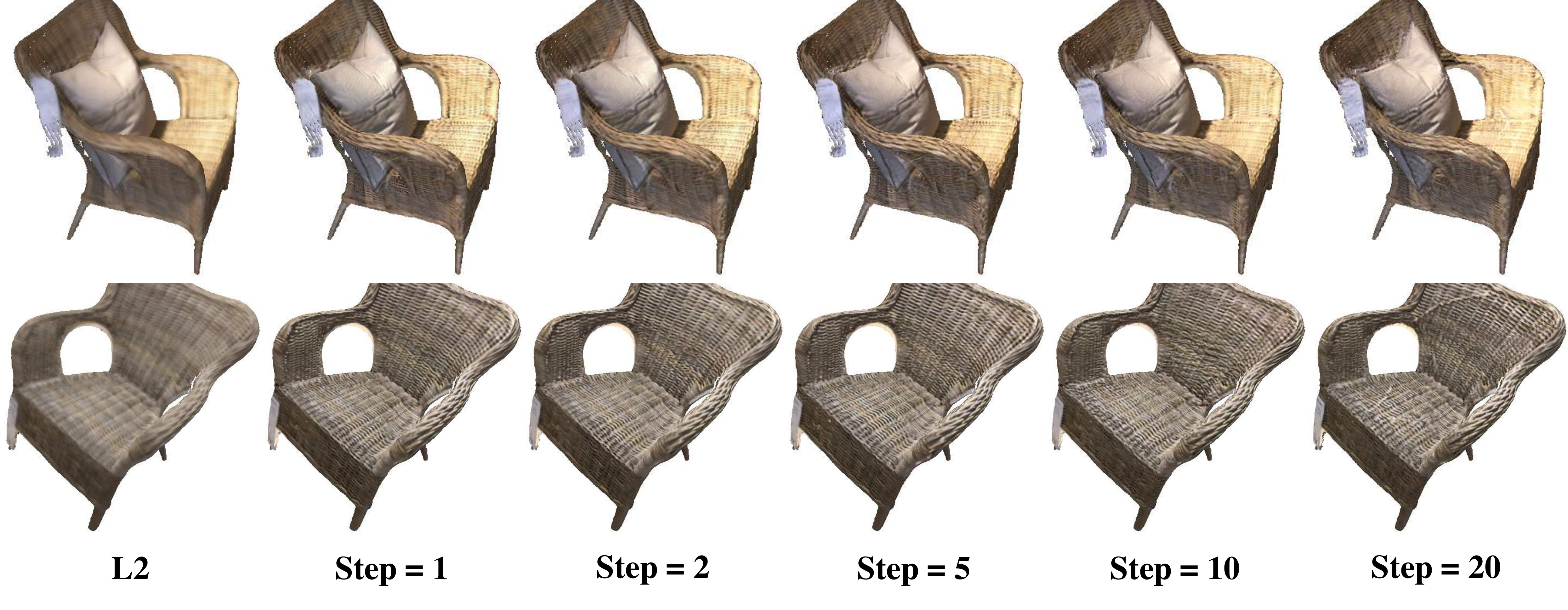}
    \caption{Ablation study on view density required for training. The two video sequences are with 426 and 244 frames respectively. The first column shows the average of frames under L2 loss. We sample the sequence by uniformly pick frames every k steps, with $k\in\{1,2,5,10,20\}$ in the remaining five columns.}
    \label{fig:sparse-view}
\end{figure*}
\section{Misalignment in the Data}
\begin{figure}
    \centering
    \includegraphics[width=\linewidth]{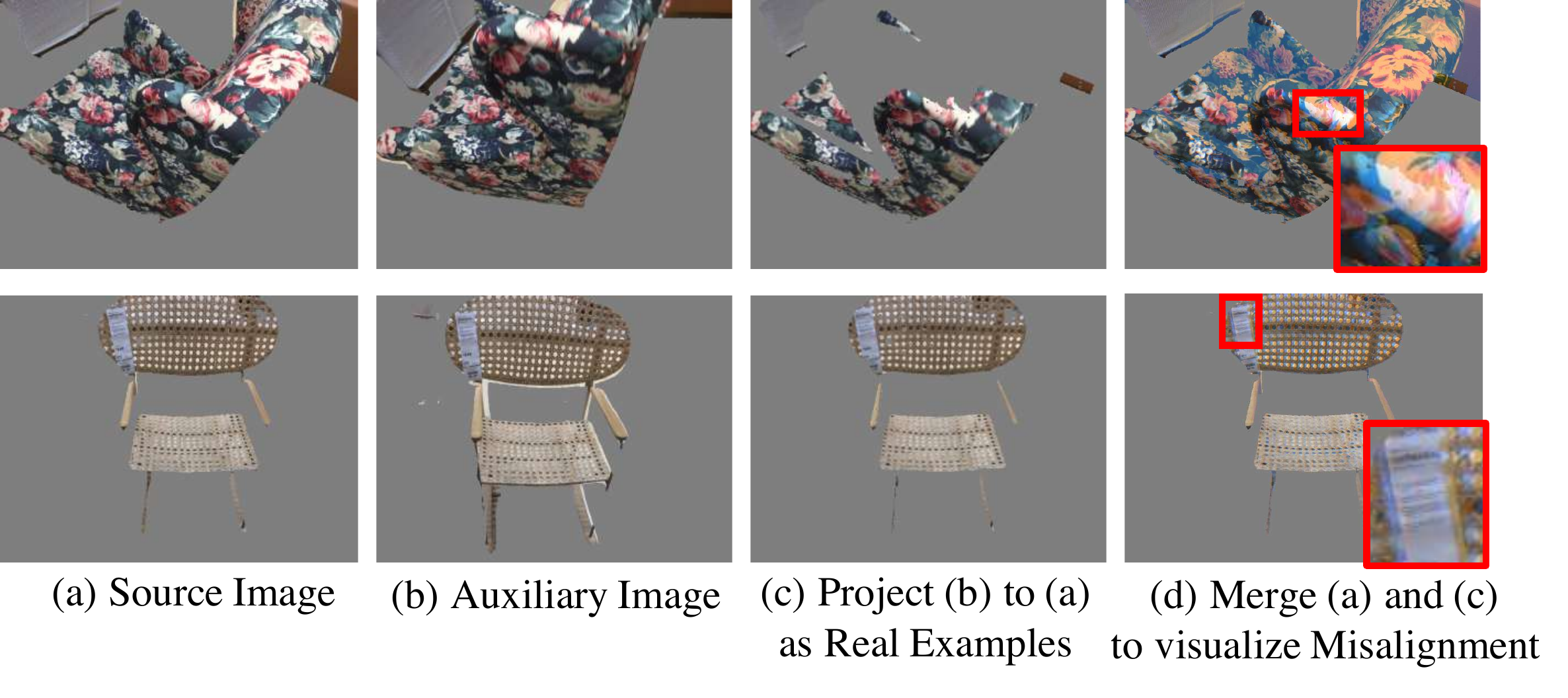}
    \caption{Real examples with Misalignment. (a) is the source image. (b) is an auxiliary image. In (c), we project (b) to the geometry and render to (a), which is a misaligned version of (a). (d) visualizes the misalignment by overlaying (a) and (c).}
    \label{fig:misalignment}
\end{figure}
Figure~\ref{fig:misalignment} shows an real example of the misaligned version of source views that we created.

\section{Sparsity of Views}
We run our algorithm on scans with different number of frames to study the behavior and the robustness of our algorithm under different level of view sparsity, as shown in Figure~\ref{fig:sparse-view}. The two video sequences are with 426 and 244 frames respectively. The first column shows the average of frames under L2 loss. We sample the sequence by uniformly pick frames every k steps, with $k\in\{1,2,5,10,20\}$ in the remaining five columns. We notice that our algorithm produces appealing results if number of frames is larger than 25, but starts to show artifact patterns under this number. We believe the reason is that with very sparse views, there are not enough patches for learning a good misalignment tolerant metric. We believe this can be addressed by data augmentation with virtual camera perturbations.

\section{Additional Comparisons}
\begin{wrapfigure}{R}{0.3\linewidth}
\includegraphics[width=\linewidth]{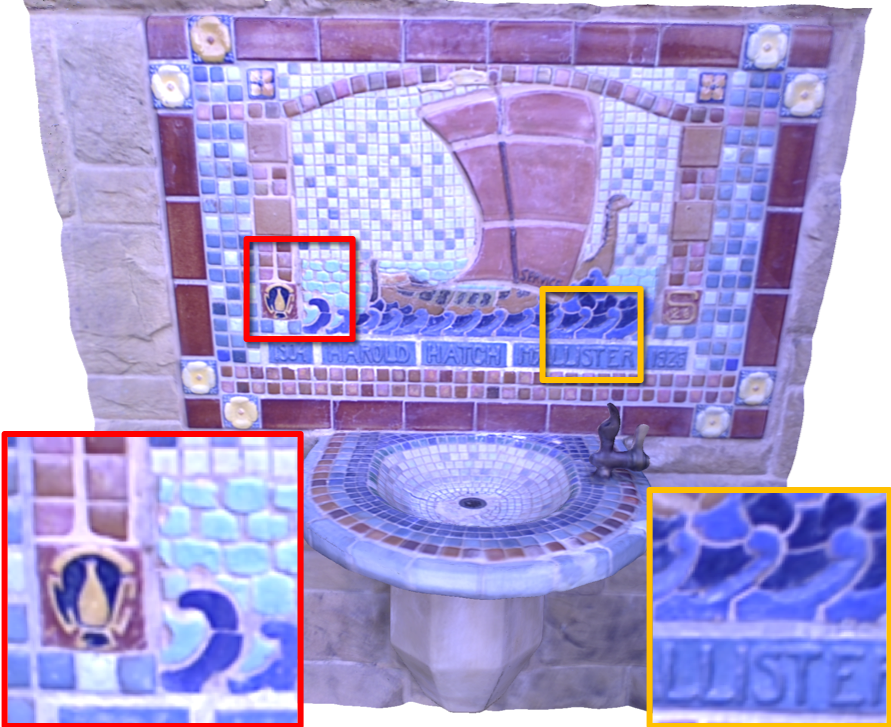}
\end{wrapfigure}
Our method can deal with standard texture optimization dataset well as shown in the example of the fountain scan. Our experiments are aimed to showcase more challenging scenarios with complex scene geometry and lighting, as well as  approximate surface reconstruction and alignment.

\begin{figure}
    \centering
    \includegraphics[width=\linewidth]{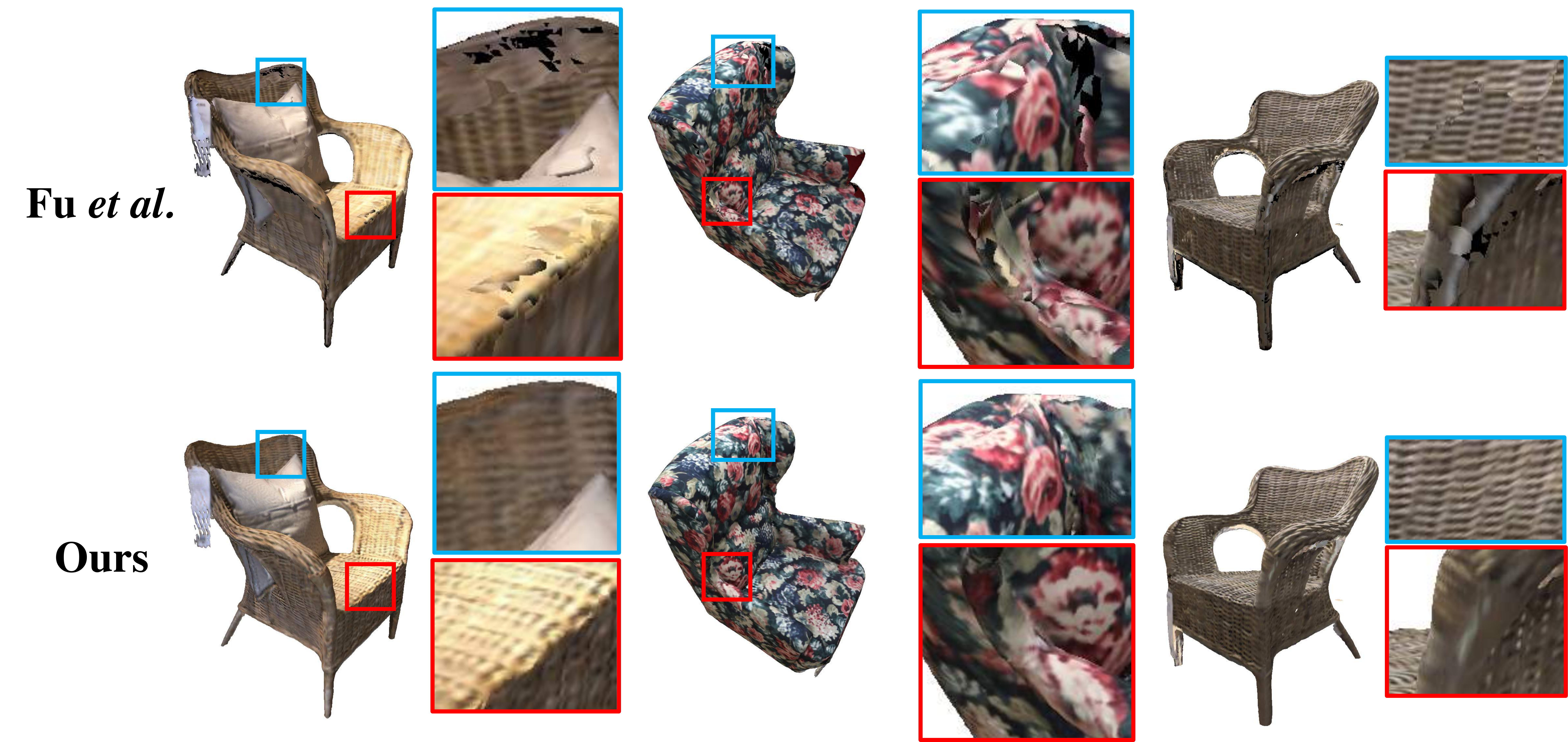}
    \caption{Comparison with~\cite{fu2018texture} that develops based on~\cite{waechter2014let}.}
    \label{fig:face-texture}
\end{figure}
We provide additional comparisons between our method and Fu \textit{et al.}~\cite{fu2018texture} that develops based on~\cite{waechter2014let}. View selection-based method yields inconsistent boundaries, while our method generates consistent texture.

 \section{Additional Results}
 We provide a video called ``supp/video.mp4'' that contains the explanation of our approach and part of video and image results.
 We additionally provide the visualization of the full real dataset in our experiments with 200 renderings comparing to different methods for objects, scenes and CAD models.
 Please check ''supp/*.html'' for details.
 
 \end{appendix}